\documentclass[runningheads]{llncs}
\usepackage{graphicx}
\usepackage{wrapfig}
\usepackage{enumitem}
\newcommand{\approptoinn}[2]{\mathrel{\vcenter{
  \offinterlineskip\halign{\hfil$##$\cr
    #1\propto\cr\noalign{\kern2pt}#1\sim\cr\noalign{\kern-2pt}}}}}
\newcommand{\appropto}{\mathpalette\approptoinn\relax}
\begin{document}
\title{Inferring and Conveying Intentionality:\\ Beyond Numerical Rewards to Logical Intentions }
\titlerunning{Inferring and Conveying Intentionality}
\author{Susmit Jha \and John Rushby}
\authorrunning{Jha and Rushby}
\institute{Computer Science Laboratory\\
SRI International\\Menlo Park CA 94025 USA\\
\email{susmit.jha@sri.com, rushby@csl.sri.com}
}
\setlength{\textheight}{19.9cm}
\maketitle              
\vspace*{-3ex}
\begin{abstract}
Shared intentionality is a critical component in developing
conscious AI agents capable of collaboration, self-reflection, deliberation, and
reasoning. 
We formulate inference of shared intentionality as an inverse 
reinforcement learning problem with logical reward
specifications.  We show how the approach can 
infer task descriptions from demonstrations. 
We also extend our approach 
to actively convey intentionality.
We demonstrate the approach on a simple 
grid-world example.
\vspace*{-3ex}
\end{abstract}
{\small Presented at AAAI Spring Symposium: Towards Conscious AI Systems, March 2019}
\section{Introduction}

There are many theories of consciousness; most
propose some biological or other mechanism as a cause or correlate of
consciousness, but do not explain what consciousness is for, nor what
it does \cite{Rushby&Sanchez18}.  We take the contrary approach: we postulate that
consciousness implements or is associated with a fundamental aspect of
human behavior, and then we ask what mechanisms could deliver this
capability and what AI approximations might help explore and validate
(or refute) this speculation.

We postulate that \emph{shared intentionality}
\cite{Tomasello-etal:intentions05} is the attribute of human cognition
whose realization requires consciousness.  Shared intentionality is
the ability of humans to engage in teamwork with shared goals and
plans.  There is no doubt that the unconscious mind is able to
generate novel and complex goals and plans; the interesting question
is how are these communicated from the mind of one individual (let's
call her Alice) to those of others so that all can engage in
purposeful collaboration.  The goal or plan is generated by some
configuration of chemical and electrical potentials in Alice's
neurophysiology and one possibility is that salient aspects of these
are abstracted to yield a concise explanation or description that
Alice can communicate to others by demonstration, mime, or language.
The description is received by the other participants (let's call the
prototypical one Bob) who can then interpret or ``concretize'' it to
enrich their own unconscious neurophysiological configuration so that
it is now likely to generate behaviors that advance the common goal.

This account suggests a dual-process cognitive architecture
\cite{Evans&Stanovich13,Frankish10,Kahneman11} where we identify
consciousness with the upper level (``System 2'') that operates on
abstracted representations of salient aspects of the lower,
unconscious level (``System 1'').  It can also be seen as a form of
Higher-Order Thought (HOT, that is thoughts about thoughts) and thus
related to HOT theories of consciousness \cite{Gennaro04}.

We posit that the conscious level is concerned with the construction
and exploitation of shared intentionality: it generates, interprets,
and communicates succinct descriptions and explanations about shared
goals and plans.  For succinctness, it operates on abstracted
entities---symbols or concepts---and presumably has some ability to
manipulate and reason about these.  When Alice builds a description to
communicate to Bob, she must consider his state of knowledge and point
of view, and we might suppose that this ``theory of mind'' is
represented in her consciousness and parameterizes her communication.

We noted that Alice could communicate to Bob by demonstration, mime
(i.e., demonstration over symbols), or language.  For the latter two,
Alice must have the abstracted description in her consciousness, but
it is possible that demonstration could be driven directly by her
unconscious: we have surely all heard or said ``I cannot explain it,
but I can show you how to do it.''  In fact, it could be that 
Alice constructs her abstraction  by mentally demonstrating the task
to herself.

In this paper, we focus on demonstration as a means for communication
and construction of abstract descriptions.  In particular, we
investigate how AI agents could use demonstrations to construct
approximations to shared intentionality that allow them to engage in
teamwork with humans or other AI agents, and to understand the
activities of their own lower-level cognitive mechanisms.

The computer science topic that seems most closely related to the task
of inferring intentionality is inverse reinforcement learning (IRL).
In classical IRL, the objective is to learn the reward function
underlying the (System 1) behavior exhibited in the demonstrations.
Here, we employ an extension to IRL that infers logical specifications
that can enable self-reflective analysis of learned
information, compositional reasoning, and integration of learned
knowledge, which enable the System 2 functions of a conscious AI
agent.  

While modern deep learning methods \cite{deeplearn} show great promise
in building AI agents with human-level System 1 cognitive capabilities
for some tasks \cite{szegedy-CVPR15,taigman-CVPR14}, and decades of
research in automated reasoning~\cite{robinson-handbook01} can be
exploited for logical deduction in System 2, our goal is to bridge
these levels by inferring and conveying logical intentions.  In this
paper, we build on previous work on logical specification mining,
including our own recent
work~\cite{marcell-nips18,jha-rv17,jha-jar18}.  The key novel
contributions of this paper are:
\begin{itemize}
    \item Formulating  intentionality inference as IRL with logical reward specification. 
    \item Methods for actively seeking and conveying intentions.
    \item Demonstration of the proposed approach on a simple grid-world example.
\end{itemize}

In Section 2,
we formulate the problem of inferring intentionality as an inverse
reinforcement learning problem and point out the deficiencies of using
numerical rewards to represent intentions.  In Section 3,
we present an inverse reinforcement learning method for logical
specifications, and illustrate how it can be used to infer intentionality.
We extend our approach to convey intentionality interactively 
in Section 4, 
and conclude in Section 5 by discussing the current limitations.

\section{IRL and Intentionality Inference}
In traditional 
Inverse Reinforcement Learning (IRL) \cite{irl}, there is a learner and
a demonstrator. The demonstrator operates in a
stochastic environment (e.g., a Markov Decision Process), and is assumed to attempt to (approximately)
optimize some unknown reward function
over its behavior trajectories.
The learner 
attempts to reverse engineer this reward
function from the demonstrations. 
This problem of learning rewards
from the demonstrations can
be cast as a Bayesian inference
problem~\cite{bayesian-irl} to predict the most probable
reward function. Ideally, this reward function encodes
the intentionality of the demonstrator and enables
the observer to understand the goal behind
the demonstrations. 

\begin{figure}
	\centering
	\includegraphics[width=4in]{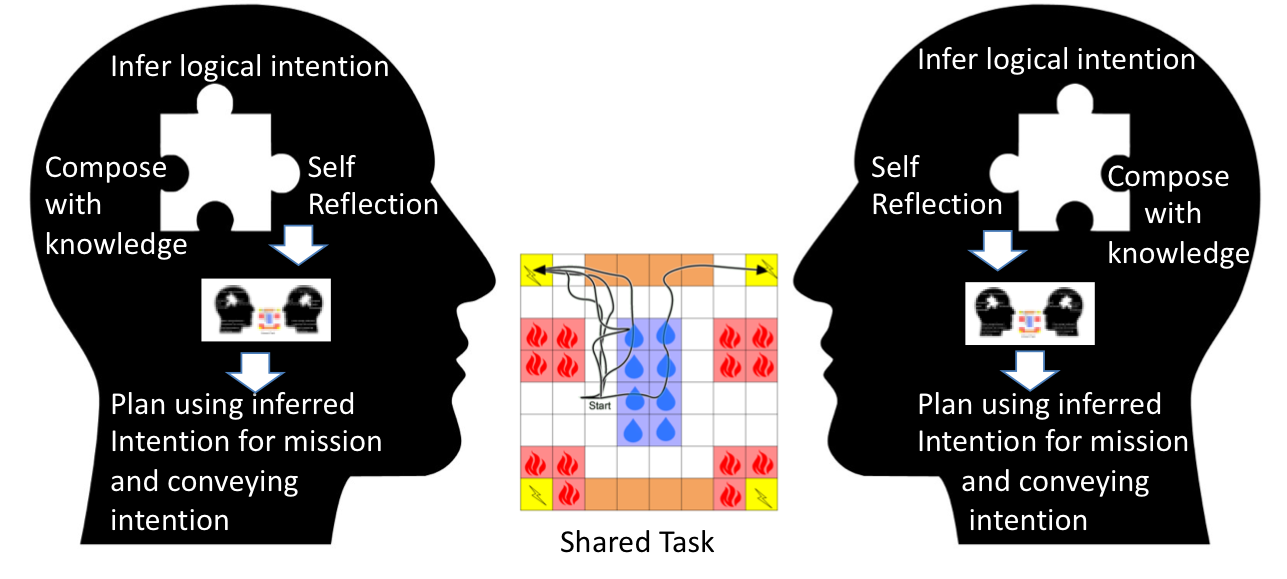}
	\caption{ AI Agents Using Intentionality Inference for Planning and Collaboration: Agents observe demonstrated behavior trajectories to formulate logical specifications that can be composed with existing knowledge about self and environment to plan out further behavior. This planning takes into account an agent's understanding of the intentions of other agents, and can be used to convey its own intentions or seek clarification about the intentions of other agents.}
	\label{fig:main}
\end{figure}

This classical form of IRL can be seen as a communication at Level 1:
that is, of an opaque low-level representation.  We enrich
this communication to allow inference of reasoning-friendly
representations such as logical specifications that are suitable for
Level 2 manipulation.  Once the agent has learned the goal in this
form, it can use its own higher-level skills and knowledge to achieve
or contribute to the goal, either independently or composed with other
goals.  Further, the agent also can use this representation to
collaborate and plan activities with other agents as illustrated in
Figure~\ref{fig:main}.

\vspace{-0.5cm}
\vspace{-0.5cm}
Logical specification mining has been studied in the traditional 
formal methods community~\cite{specmining} including our 
own past work~\cite{jha-rv17,jha-nfm17,jha-jar18}, but these methods are not
robust to noise and rely on intelligent oracles to produce behaviors that cover the space of legal behaviors for the specification. This is not 
realistic for general AI problems where demonstrations such as handing over a glass of water, or crossing a street, are inherently noisy. 
In contrast, 
IRL algorithms~\cite{maxentropyIRL} 
formulate this inference procedure using the principle of 
maximum entropy~\cite{motherofall}. This results in a likelihood
of inferred reward 
over the demonstrations which is no more committed to any particular
behavior than is required for matching the empirically observed
reward expectation. 
Traditionally, this approach was limited to structured scalar rewards,
often assumed to be linear combinations of feature vectors. But more 
recently, these have been adapted to arbitrary function 
approximators such as
Gaussian processes~\cite{levine-gaussianIRL} and neural
networks~\cite{finn-nnIRL}.
While powerful, these existing IRL methods provide 
no principled mechanism
for composing or reasoning with the resulting rewards. The 
inference of intention as numerical reward function 
lacks a form that is amenable for self-reflection
and collaboration, and has several limitations:
\begin{itemize}

\item First, numerical reward functions lack logical structure, making
it difficult to reason over them---which is critical for
self-reflection: a conscious AI agent must be able to analyze its
understanding of intention.   This inference of intention could be
from behaviors (either real or mental rehearsals) of
its own low-level cognitive system, or from behaviors of
other conscious agents.

\item Second, combining numerical rewards to understand intention in a
compositional manner is difficult.  Demonstrations for two tasks can
be learned individually using numerical rewards but these cannot be
combined by the AI agent to perform the tasks in a concurrent or
coordinated manner.  A conscious AI agent cannot just infer each
task's intention separately, but needs a global view of its own
inference and understanding.

\end{itemize}

\section{IRL with Logical Intention Discovery}
In this section, we briefly summarize how our recent work~\cite{marcell-nips18}
on inferring
logical specifications in IRL can be used to  answer the foundational Question 1 stated below. This is the first step required to build self-aware and self-reflective AI agents capable of inferring and 
conveying intentions. 
\begin{quote}
    \textbf{Question 1.} \emph{How does Alice infer logical specification of intention by observing a set of demonstrative behaviors (either Alice's own behavior generated by lower-level cognitive engines, or that of another agent)?}
\end{quote}

We assume that the demonstrator (Alice or Bob) operates within
a Markov Decision Process and the specification of the intent is
a bounded trace property. More precisely, we define a
demonstration/trajectory, $\xi$, to be a sequence of
state-action pairs. Alice attempts to infer past-time linear 
temporal logic (PLTL)~\cite{havelund-STTT04}
from the demonstrations. 
Such a PLTL
property, $\phi$, can be identified as a binary non-Markovian
reward function $\phi: \xi \rightarrow 1 \;$ if $\; \xi \models \phi$, and $0$ otherwise. The candidate set of 
specifications corresponding to the space of 
possible intentions is denoted by $\Phi$. Inferring
intention from demonstrations in the set $X$ 
can be formulated
as a maximum posterior probability inference problem:
$ \phi^* = \arg \max_{\phi \in \Phi} Pr(\phi | X) $.
Under assumptions of uniform prior over the intention 
space, and applying maximum entropy principle (see \cite{marcell-nips18} for technical details), the posterior probability of a specification is given by:
$$ Pr (\phi | M, X, \overline \phi) \appropto \mathbf{1}[\overline \phi \geq \hat \phi] \cdot \exp \big ( |X| \cdot D_{KL} ( \mathcal{B}(\overline \phi) || \mathcal{B}(\hat\phi) ) \big )$$
where $M$ is the stochastic dynamics model known to the agent, $X$ is the set of demonstrations, $\overline \phi$ denotes the average number of times the specification $\phi$ was satisfied by the demonstrations, $\hat \phi$ denotes the average number of times the specification is satisfied by a random sequence of actions, and $D_{KL}$ denotes the KL divergence between the two Bernoulli distributions denoted by
$\mathcal B$. 
Intuitively, the first component is an indicator function that the demonstrator is better than random, 
and the second component measures the information gain 
over the random actions. 
We can obtain the most likely logical specification from a set of demonstrations by maximizing the posterior probability. An algorithm
for this optimization using partitioning of the logical specifications is presented in our 
previous work~\cite{marcell-nips18}.

\begin{wrapfigure}{r}{0.3\textwidth} 
\vspace{-20pt}
  \begin{center}
    \includegraphics[width=0.29\textwidth]{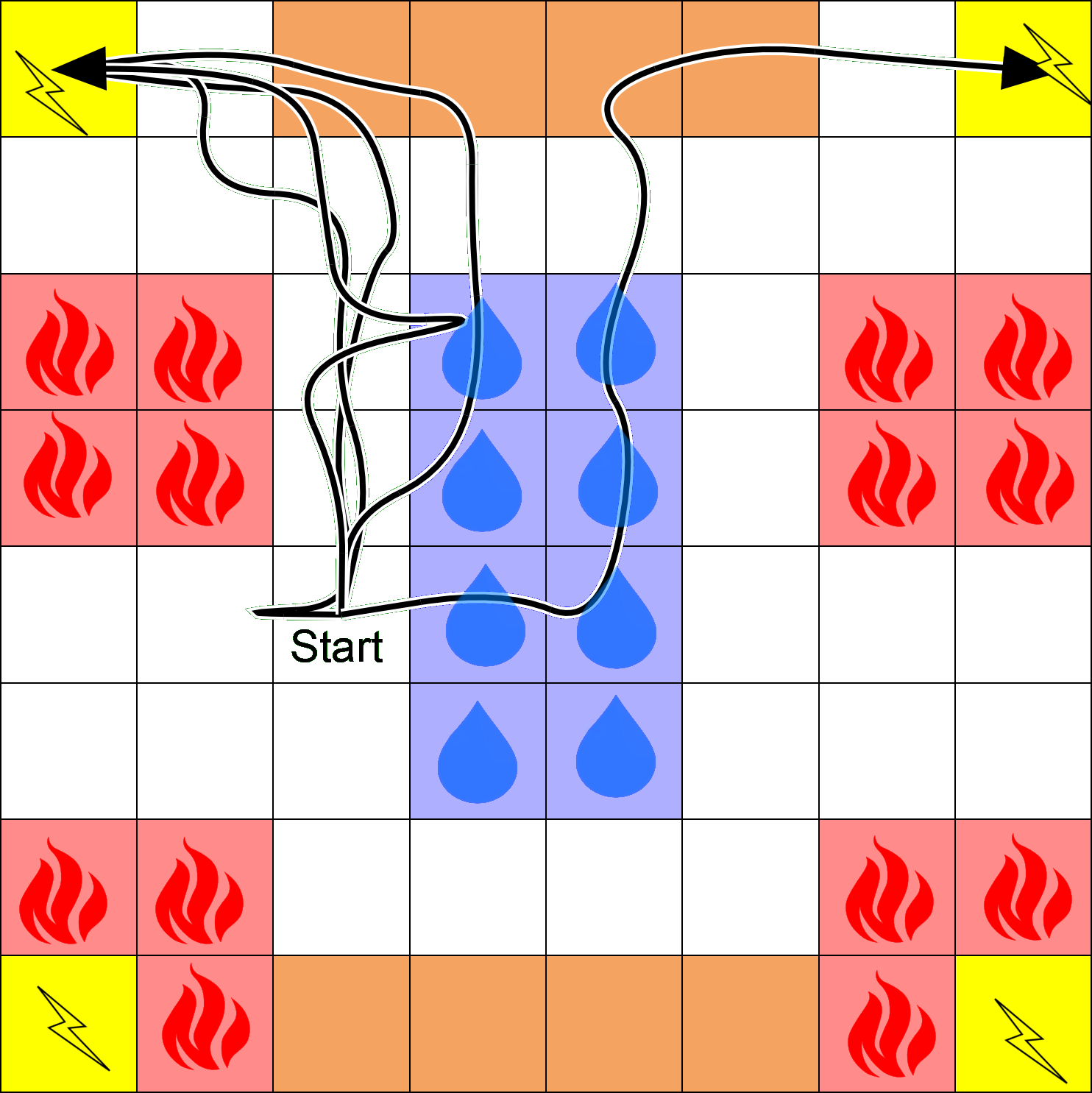}
    \caption{\label{fig:gridworld}}
  \end{center}
  \vspace{-20pt}
  \vspace{1pt}
\end{wrapfigure} 
We use a simple grid world example to demonstrate this
approach illustrated in 
Figure~\ref{fig:gridworld}. 
In this task, the agent moves in a discrete gridworld and can take actions to move in the cardinal directions (north, south, east, west). Further, the agent can sense abstract features of the domain represented as colors. The task is to reach any of the yellow (recharge) tiles without touching a red tile (lava) -- we refer to this sub-task as YR. Additionally, if a blue tile (water) is stepped on, the agent must step on a brown tile (drying tile) before going to a yellow tile -- we refer to this sub-task as BBY.  The last constraint requires recall of two state bits of history (and is thus not Markovian and infeasible to learn using traditional IRL): one bit for whether the robot is wet and another bit encoding if the robot recharged while wet. Demonstrations correspond to simultaneously satisfying both requirements. 
The space of logical specifications~\cite{temporallogic} 
consist of 
PLTL properties using atomic propositions that indicate the nature of the square on which the robot is at a given instant.  These demonstrations are interesting because they incidentally include noisy
demonstrations for incorrect intentions, for instance, the robot should wet and dry itself before charging. But our algorithm using max entropy principle infers the following correct requirement using
approximately 95 seconds and after exploration of  172 $\hat{\varphi}$
candidates ($\approx 18\%$ of the concept class):
$ \phi_F \equiv \big ( H \neg red \wedge O \; yellow \big ) \wedge H \big ( (yellow \wedge O\; blue) \Rightarrow (\neg blue \;S\; brown ) \big )$,
where $H$ is ``historically,'' $O$ is ``once,'' and $S$ is ``since'' ~\cite{havelund-STTT04}.

\section{Passive Inference to Active Transfer of Intention}
A conscious agent must be capable of active transfer of intention
beyond passive inference of intent discussed above. Such active intent transfer includes: 
\begin{quote}
    \textbf{Question 2.} \emph{How does Alice infer (and
    then correct) a gap in the logical specification of her  intention learned by Bob ?}
\end{quote}
\begin{quote}
    \textbf{Question 3.} \emph{How does Alice seek clarifying behaviors from Bob to disambiguate her currently inferred intentions of Bob ?}
\end{quote}

The key to addressing 
both questions lies in defining a divergence 
measure over the set of candidate specifications representing
possible intention. One such divergence measure is the ratio
of log likelihoods of two specifications $\phi$ and $\phi'$: 
\begin{eqnarray*} 
D(\phi,\phi') & = & \log(  Pr (\phi | M, X, \overline \phi)  /   Pr (\phi' | M, X, \overline \phi') ) \\
& = & D_{KL} ( \mathcal{B}(\overline \phi) || \mathcal{B}(\hat\phi) ) -  D_{KL} ( \mathcal{B}(\overline \phi') || \mathcal{B}(\hat\phi') )
\end{eqnarray*}
We also assume both Alice and Bob have common intent inference mechanism which allows them to run the algorithm over demonstrations, and infer what the other agent might be concluding so far. Extension of this approach to agents who use different background knowledge, and will have noisy simulation of the other agent's intention inference mechanism is beyond the scope of this paper.

To demonstrate the use of this divergence measure, we consider a scenario where the demonstrations on the grid-world 
are restricted to a subset $X'$ of original set $X$, and 
$X'$ does not contain any trajectories going through blue or brown tiles. Using these demonstrations, Alice 
 infers  
$\phi_{YR} \equiv H \neg red \wedge O \; yellow  $
as the most likely explanation, which only corresponds to the sub-task of avoiding lava and reaching the recharge tile. Alice can evaluate other specifications and, if there are other candidate 
specifications with low divergence measure, she can 
attempt to disambiguate her inferred intent. 
Let us say one such specification is 
$\phi \equiv H \neg red \wedge O \; yellow \wedge O \; blue  $. Alice can generate demonstrations consistent with this specification
by planning from temporal logic~\cite{jha-jar18b}. 
These demonstrators will pass through wet blue tiles,
and reach recharge without visiting brown drying tiles. 
Bob runs the intent inference approach on these demonstrations to realize that Alice has inferred $\phi$, and not the intended 
$\phi_{YR}$. He can provide additional behaviors (for e.g., the original set $|X|$) that help disambiguate both specifications. This is continued until Alice converges to $\phi_F$, and all other candidate specifications having high divergence from $\phi_F$. 

\section{Conclusion}
In this paper, we presented a first step towards building AI agents
capable of inferring and conveying intentionality as logical specifications.
The goal is to develop AI agents that not only learn intentions of 
other agents from demonstrations, or their own intentions by observing 
actions of lower-level cognitive engines, but also to provide and seek
clarifications when inferred intentions are ambiguous. Our proposed
approach is currently
limited to behaviors which are represented as time traces,
and intentions that can be expressed in temporal logic. But several 
creative tasks
such as proving theorems or writing a mystery novel cannot be
easily formulated in this framework. 
A hierarchical 
representation mechanism that can exploit the inferred intentions
and goals to compositionally learn new intentions is essential to 
building self-aware self-reflective AI that can collaborate to 
perform creative endeavors. 
\vspace{-0.3cm}
\paragraph{Acknowledgement:}
The  authors  acknowledge  support  from  the  National  Science  Foundation(NSF) Cyber-Physical Systems \#1740079 project, NSF Software \& Hardware Foundation \#1750009 project, and US ARL Cooperative Agreement W911NF-17-2-0196 on Internet of Battle Things (IoBT).

\bibliographystyle{splncs04}

\end{document}